\documentclass{article}
\usepackage{spconf,amsmath,graphicx,hyperref,amsfonts}
\usepackage[sort&compress,numbers,square]{natbib}
\usepackage{booktabs}
\usepackage{multirow}
\usepackage{makecell}
\usepackage{xcolor}
\newcommand{\methodname}{InfoGain Wavelets}
\newcommand{\methodnameshort}{InfoGain}
\newcommand{\methodnamecaps}{INFOGAIN WAVELETS}

\title{{\methodnamecaps}: Furthering the Design of  Graph Diffusion Wavelets}

\threeauthors
  {David R. Johnson\thanks{D.J., S.K., and M.P. were supported in part by NSF DMS Award \#2327211.}}{Boise State University\\Program in Computing\\Boise, ID, USA}
  {Smita Krishnaswamy}{Yale University\\Department of Computer Science\\Department of Genetics\\New Haven, CT, USA}
  {Michael Perlmutter}{Boise State University\\Department of Mathematics\\Program in Computing\\Boise, ID, USA}
  
\begin{document}
%\ninept
%
\maketitle
\begin{abstract}
Diffusion wavelets extract information from graph signals at different scales of resolution by utilizing graph diffusion operators raised to various powers, known as diffusion scales. Traditionally, these %diffusion
scales are chosen to be dyadic integers, $2^j$. Here, we propose a novel, unsupervised method for selecting the diffusion scales based on ideas from information theory. We then show that our method can be incorporated into wavelet-based GNNs, which are modeled after the geometric scattering transform, via graph classification experiments. 
\end{abstract}
\begin{keywords}
Wavelets, Geometric deep learning, Graph neural networks, Geometric scattering transforms 
\end{keywords}
\section{Introduction}
\label{sec:intro}

%%%%%%%%%%%%%%%%%%%%
%   INSTRUCTIONS   %
%%%%%%%%%%%%%%%%%%%%
% \textcolor{red}{``You are allowed a total of 5 pages for your document. Up to 4 pages may contain technical content, figures, and references, while the 5th page may contain only references, funding acknowledgments, and a Compliance with Ethical Standards statement.`` Throughout, we need to replace uses of \texttt{citet} with manually entering the author(s)' names followed by cite, since the style file provided is ancient and requires this.}

Diffusion wavelets were introduced in \cite{coifman:diffWavelets2006} by Coifman and Maggioni in order to extend wavelets analysis to geometric domains such as graphs and manifolds. Analogous to traditional wavelets  used for processing Euclidean data such as images \cite{Mallat:2008:WTS:1525499}, diffusion wavelets aim to capture information about the input signal at multiple scales of resolution. From the perspective of graph signal processing \cite{shuman:emerging2013}, they can be thought of as band-pass filters, where each wavelet filter highlights a different frequency band.

%\textcolor{red}{Potential strategy to reduce references count: move the bulk of introduction with heavy references into the `Background and motivation` or a (new) `Related work` section in the appendix?} 
%%% NOTED
% Commenting out for now to help with editting
%%%
Subsequently, as part of the rise of geometric deep learning \cite{Bronstein:geoDeepLearn2017,bronstein2021geometric}, several authors have used diffusion wavelets as the basis for geometric scattering transforms \cite{gama:diffScatGraphs2018,gama:stabilityGraphScat2019,gao:graphScat2018,zou2020graph, perlmutter:geoScatCompactManifold2020,chew2022geometric,bodmann2022scattering,pan2020spatio,ioannidis2020efficient} (GSTs), modeled after a similar construction introduced by Mallat for Euclidean data \cite{mallat:scattering2012} (see also \cite{czaja:timeFreqScat2017, nicola2022stability, bruna2013invariant, grohs:cnnCartoonFcns2016, wiatowski:mathTheoryCNN2018}). The GST operates similarly to a convolutional neural network, in that each input signal $\mathbf{x}$ undergoes an alternating sequence of filter convolutions and pointwise nonlinearities (and then possibly a final low-pass filter or global aggregation). The output of these operations, referred to as \emph{scattering coefficients}, can then be used as input to a regressor or a classifier. %, or other machine learning algorithm. %machine learning algorithm of the users choice. 
%%%%
% MP: Can also use scattering for clustering, e.g. k-means on top of scattering coefficients
%%%%%

Notably, the initial versions of the GST, as well as Mallat's Euclidean scattering transform, differ from standard deep feed-forward architectures in that they were \emph{predesigned}. This facilitates the theoretical analysis of such networks 
\cite{gama:diffScatGraphs2018,gama:stabilityGraphScat2019,perlmutter:geoScatManifolds2018,perlmutter2019understanding,mallat:scattering2012,nicola2022stability} 
and also makes  them naturally well suited towards unsupervised learning or low-data environments \cite{saito2017underwater,LEONARDUZZI201811}. However, subsequent work, \cite{wenkel2022overcoming,tong2022learnable,xu2023blis}, has shown that diffusion wavelets can also be incorporated into fully learned Graph Neural Networks (GNNs), which may be thought of as learnable versions of the scattering transform. They have shown that these wavelet-based GNNs are effective for overcoming the oversmoothing problem \cite{wenkel2022overcoming} (via the use of band-pass filters as well as low-pass) and for solving combinatorial optimization problems \cite{min2020scattering, wenkeltowards}. The purpose of this paper is to further the design of diffusion wavelets (and thus associated GSTs and GNNs) with a novel, unsupervised approach based on information theory. %the Kullback-Leibler (KL) divergence.

More specifically, diffusion wavelets aim to extract information at multiple levels of resolution by considering various powers of a diffusion matrix such as the lazy random-walk matrix $\mathbf{P}$. Traditionally, these powers, referred to as \emph{diffusion scales}, are chosen to be powers of two, a choice inherited from Euclidean wavelets. In the context of images, these dyadic scales are quite natural since the domain of the signal, the unit-square $[0,1]^2$, may naturally be divided into squares of length $2^j$ for differing values of $j$. However, in the context of data with irregular geometric structure, such as graphs, this choice is somewhat less natural and may limit performance. Indeed, this observation was the basis for \cite{tong2022learnable}, which proposed to learn the optimal diffusion scales via a differentiable selector matrix. Here, we propose a different, information-theoretic approach for designing diffusion scales. Notably, unlike \cite{tong2022learnable}, our approach is unsupervised 
%making it easier to apply in low-data environments \textcolor{orange}{avoiding this for now, since (a) we don't do experiments explicitly testing this versus LEGS, and (b) the original LEGS paper does do experiments showing its low-data prowess}. Additionally, our method differs from \cite{tong2022learnable} in that it 
and can learn a separate set of scales for each input signal.  

%Diffusion Wavelets If space some description of wavelets in signal processing

\section{Background}\label{sec: background}
% Let $G=(V,E,w)$ be an weighted, undirected graph with weighted adjacency matrix {$\mathbf{A}$ and weighted degree matrix $\mathbf{D}$. Let $\mathbf{x}:V\rightarrow\mathbf{R}$ denote a signal (function) defined on the vertices of $G$ (identified with a vector), and let $\mathbf{P}=\frac{1}{2}(\mathbf{I}+\mathbf{A}\mathbf{D}^{-1})$ be the lazy random matrix. The dyadic wavelet transform of $\mathbf{x}$ is defined by $\{\Psi_j\mathbf{x}\}_{j=0}^J\cup\{\Phi_J\mathbf{x}\}$, where $\Psi_0=\mathbf{I}-\mathbf{P}$, $\boldsymbol{\Phi_J}=\mathbf{P}^{2^J}$, and $\Psi_j=\mathbf{P}^{2^{j-1}} - \mathbf{P}^{2^{j}}$.
Let $G=(V,E,w)$ be an weighted, undirected graph, $|V|=n$, with weighted adjacency and degree matrices {$\mathbf{A}$ and $\mathbf{D}$,  and let $\mathbf{P}=\frac{1}{2}(\mathbf{I}+\mathbf{A}\mathbf{D}^{-1})$ be the lazy random matrix. For a graph signal (function) $\mathbf{x}:V\rightarrow\mathbb{R}$ (identified with a vector in $\mathbb{R}^n$), we define its \emph{dyadic wavelet transform}  by $\mathcal{W}_J\mathbf{x}=\{\Psi_j\mathbf{x}\}_{j=0}^J\cup\{\Phi_J\mathbf{x}\}$, where $\Psi_0=\mathbf{I}-\mathbf{P}$, $\Phi_J=\mathbf{P}^{2^J}$, and 
\begin{equation}
\label{eqn: dyadic coefficients}\Psi_j=\mathbf{P}^{2^{j-1}}- \mathbf{P}^{2^j},\quad\text{ for }1\leq j \leq J,
\end{equation}
where $J\geq 0$ is a hyperparameter.

%\IEEEpubidadjcol
The geometric scattering transform is a multilayer, nonlinear feed-forward architecture based on the wavelet transform. For each input signal $\mathbf{x}$, it defines first- and second-order scattering coefficients\footnote{Higher-order coefficients can be defined by similar formulas.} by
\begin{equation}\label{eqn: scat}
\mathcal{U}[j]\mathbf{x}=\sigma(\Psi_j\mathbf{x}) \text{ and } \mathcal{U}[j_1,j_2]\mathbf{x}=\sigma(\Psi_{j_2}\sigma(\Psi_{j_2}\mathbf{x})),
\end{equation}
where $\sigma$ is a pointwise activation operator. %, the modulus operator $M\mathbf{x}(v)=|\mathbf{x}(v)|$. 
We note that the use of alternating linear transformations (wavelet filterings) and non-linear activations (the modulus operator) is meant to mimic the early layers of a neural network. When used for supervised learning, one may treat these coefficients as new features to be fed into a downstream learning algorithm.%After computing the scattering coefficients, one can them feed them into another machine learning architecture such as a multi-layer perceptron.
%\IEEEpubidadjcol

Importantly, we note that the wavelet transform, and therefore also the scattering coefficients, can be computed efficiently via recursive sparse matrix-vector multiplications. In particular, one never needs to form a dense matrix. This allows for the geometric scattering transform, and related GNNs \cite{wenkel2022overcoming}, to be applied on large networks  such as those found in the open graph benchmark data sets \cite{hu2020open}.

Part of the utility of the geometric scattering transform is that it allows one to avoid the oversmoothing-versus-underreaching tradeoff. Standard message-passing neural networks rely on localized averaging type operations, which from the standpoint of graph signal processing are viewed as low-pass filters \cite{nt2019revisiting}. These filters progressively smooth the node features and so the number of layers must be kept small in order to avoid severe oversmoothing. However, this creates a new problem, underreaching. 

By design, the wavelets $\Psi_j$ have a receptive field of length $2^j$. Furthermore, these wavelets can be understood as band-pass filters rather than low-pass. Therefore, scattering based networks are able to capture global structure without oversmoothing. This is particularly important for molecular graphs, and other biomedical data sets, which do not exhibit the small world phenomenon. Therefore, the GST, and other wavelet based networks are particularly effective in these settings \cite{viswanath2024protscape,bhaskar2022molecular,venkat2024directed,viswanath2025hiponet,chew2022manifold,zou:graphScatGAN2019}. %Additionally

% and limit performance on some data sets
However, it has been observed that the   predesigned choice of using dyadic scales, $2^j$, may be overly rigid and hinder performance. Accordingly, \cite{tong2022learnable} proposed \emph{generalized diffusion wavelets}, with formulas similar to \eqref{eqn: dyadic coefficients}, but where the dyadic scales $2^j$ are replaced by an arbitrary sequence of scales $t_0<t_1<\ldots<t_J$. This lead to a  Learnable Geometric Scattering (LEGS), which used formulas similar to \eqref{eqn: scat}, but where the scales were learned via a differentiable selector matrix. %that aims to learn one optimal set of scales 
(For details, 
see Appendix \ref{app:legs}). In this work, we provide an alternative to LEGS for selecting the diffusion scales, motivated by the following considerations:

\begin{itemize}
\item \textbf{Unsupervised learning and low-data environments:} LEGS utilizes a differentiable selector matrix, and therefore requires labeled training data. However, previous work \cite{venkat2024mapping,sun2024hyperedge} shows that the diffusion wavelets can be effective for unsupervised learning. Is there a better way to choose diffusion scales for unsupervised tasks or low-data environments?
\item \textbf{Sparsity and efficiency:} The rows of the selector matrix are only \emph{approximately} sparse. This means that LEGS must utilize dense matrix-vector multiplications throughout  training. (Note that %since LEGS is end-to-end differentiable, 
the locations of the dominant scales may change during training.) Is there a more precise way to learn the diffusion scales that is amenable to sparse matrix operations?
\item \textbf{Channel-specific scales:} LEGS selects a single set of scales for all channels (i.e., features). What if the optimal set of scales is different for each input channel? 
\end{itemize}
    
% \textcolor{orange}{
% {\methodname} is motivated by previously unresolved questions in learnable geometric scattering networks, namely:
% \begin{enumerate}\itemsep0em
%     \item What if a graph (or manifold) data set has multiple channels of vertex (point) signal, which  have different optimal wavelet scales? LEGS aims to learn one `best' set of scales, applied to all channels.
%     \item What if a (high-dimensional) data set has too few samples for effective supervised scales learning via LEGS? (Recall that, in the original, handcrafted approach, geometric scattering is touted as effective with low-data problems \cite{gao:graphScat2018}.) Could an unsupervised approach lead to effective scale selection?
%     \item Since scale selection is a discrete learning problem (i.e. learning the indices of key discrete diffusion steps to partition a signal into bandlimited signals), is there a cheaper and/or more precise way to learn generalized wavelet scales than continuous approximation, which in LEGS means that we can't take advantage of sparse matrix/tensor operations?
% \end{enumerate}
}

%In response to the above questions on LEGS, we will contribute a new, unsupervised method of generalized diffusion wavelet scale selection for use in geometric scattering networks, entitled `{\methodname}'. 

%Explaining LEGS in more detail here makes more sense I think. We will see depending on space

\section{{\methodnamecaps}}\label{sec:algorithm}

In brief, our proposed algorithm aims to construct a sequence of diffusion scales $t^c_0<t^c_1<\ldots<t^c_J$ for each channel $c$ so that the wavelet coefficients $(\mathbf{P}^{{t^c_{j-1}}}- \mathbf{P}^{t^c_j})\mathbf{x}_c$
capture approximately equal increments of Kullback-Leibler (KL) divergence (also known as information gain or relative entropy)
%contain approximately equal amounts of information for each $0\leq j \leq J$, quantified in terms of the Kullback-Leibler (KL) divergence (also known as information gain or relative entropy) 
using $\mathbf{P}^{T_J}x$ for some large value $T_J$ as a `smooth' reference distribution. 
This allows one to construct a set of wavelets from scales
% which utilizes the most `important' scales $t_0^c<\ldots<t_J^c$ 
that mark roughly even degrees of information loss from diffusion-based signal smoothing, uniquely for each channel and without the need for large amounts of (labeled) data. 
%Further details on the motivation behind our algorithm are in Appendix \ref{app:algorithm_details}.

% We then select the remaining scales $t^c_3,\ldots,t^c_J$ based on the objective of maintaining roughly equal intervals of relative information gain (also known as relative entropy or Kullback-Leibler (KL) divergence) between each diffusion band and the $t_J$-th diffusion step.  %(We note that if $t_J$ is large, then $\mathbf{P}^{t_J}\mathbf{x}$. 
%Recall that the matrix of the diffusion operator $\mathbf{P}$ is the transition matrix of a Markov chain with a unique stationary distribution.

More formally, we fix $J\in\mathbb{N}$ and a maximal diffusion scale $t_J$. For simplicity, for each channel $c$, $1\leq c\leq C$, we always set $t^c_0=0, t^c_1=1$, and $t^c_2=2$ so that first two wavelets are always given by $\Psi_0 = \mathbf{I} - \mathbf{P}$ and $\Psi_1 = \mathbf{P}- \mathbf{P}^{2}$, the same as in the dyadic case. Then, \textit{for each graph} in the data set, %\textcolor{orange}{would an `i' subscript on the $\text{x}_{c}$s here make this more clear?}
%%%%%%%%%%%%%%%%%%%%%
% MP: I am conflicted, but think it would do more harm then good. I think it is already clear and there are already a lot of indices floating around
%%%%%%%%%%%%%%%%%%%%
we consider the $n\times C$ feature matrix $\mathbf{X}$, where each column, $\mathbf{x}_c$,  represents an input channel, %where $C$ is the number of channels, 
and:
\begin{enumerate}\itemsep0em
    \item Compute $\mathbf{P}^{t}\mathbf{x}_c$ for each $c$ and each $t = 2,\ldots,t_J$ (recursively via sparse matrix-vector multiplication).
    \item Normalize each $\mathbf{P}^{t} \mathbf{x}_{c}$ %(the lazy random walk diffusion steps of each graph and channel) 
    into probability vectors $\mathbf{q}^{t}_{c}$ for $t = 2,\ldots,t_J$ by  applying min-max scaling to map the entries %of each vector 
    into $[0,1]$ and then applying $\ell^1$-normalization.
    \item Replace the zero entries in each $\mathbf{q}^{t}_{c}$, with a minimal value, e.g., $\frac{1}{2}\mathrm{min}\{\mathbf{q}^{t}_{c}: \mathbf{q}^{t}_{c} > 0\}$ %with a small nonzero value
    to avoid arbitrarily small values from skewing information calculations. %(For example, a small value such as $10^{-2}$, or a data-derived value such as $\frac{1}{2}\mathrm{min}\{\mathbf{q}^{t}_{c}: \mathbf{q}^{t}_{c} > 0\}$.) 
    \item Compute the KL divergences %($D_{KL}$)  
    between $\mathbf{q}_c^t$ and $\mathbf{q}_c^{t_J}$ :%for $t=2,\ldots,t_{J-1}$, i.e.,
$$(D_{KL})_{t,c}=D_{KL}\big(\mathbf{q}^{t}_{c} \; \| \; \mathbf{q}^{t_J}_{c}\big) = \sum_{k=1}^{n} \mathbf{q}^{t}_{c}(k) \; \mathrm{log}\bigg(\frac{\mathbf{q}^{t}_{c}(k)}{\mathbf{q}^{t_J}_{c}(k)} \bigg).$$
\end{enumerate}

Next, \emph{for each fixed channel} $c$, $1\leq c\leq C$, we then:
\begin{enumerate}\itemsep0em
    \item Sum the $(D_{KL})_{t,c}$ over all graphs  for each time $t$: i.e., $(D_{KL})_{t,c}^{\text{total}}=\sum_{i=1}^{N_G} (D_{KL})_{t,c}(i)$, where $N_G$ is the number of graphs, and $(D_{KL})_{t,c}(i)$ is the KL divergence at time $t$ and channel $c$ on graph $i$. (We may optionally re-weight these sums for unbalanced classes, if the downstream task is graph classification. If the data consists of a single large graph, we omit this step.)
    \item  Compute  cumulative sums $S_{t,c}=\sum_{s=2}^t(D_{KL})^{\text{total}}_{t,c}$ for each $t=2,\ldots,t_{J-1}$.
    \item Apply min-max rescaling to the each of the $\{S_{t,c}\}_{t=2}^{t_{J-1}}$ so that they have minimal value $0$ and maximal value $1$. %they lie in $[0,1]$. %in order for it to have minimal value $0$ and maximal value $1$. 
    \item Select  diffusion scales $t^c_j$ %(i.e., the powers to which $\mathbf{P}$ is raised) 
    so that the $S_{t^c_j,c}$ are as evenly spaced as possible for a desired number of scales.
\end{enumerate}

After this procedure, we then define $\Psi_{j_c}=\mathbf{P}^{t_{j-1}^c}-\mathbf{P}^{t_{j}^c}$ for $2\leq j\leq J$  define $\Phi_J=\mathbf{P}^{t_J}$. We observe that by Step 4 above, the selected wavelet features $\Psi_{j_c}\mathbf{x}_c$ are approximately balanced so that each contributes a comparable share of divergence from the smooth reference distribution.
%there is approximately an equal amount of information in each of the wavelet features $\Psi_{j_c}\mathbf{x}_c$. 
Additionally, we remark that as a practical strategy for accomplishing Step 4, one can designate `selector quantiles' and pick  $t_j^c$ to be the first integer where $S_{t_{j}^c,c}$ exceeds a given quantile. For example, if one chooses selector quantiles of $[0.2, 0.4, 0.6, 0.8]$, for a channel $c$, then $t^c_{3}$ will be diffusion scale where the channel's normalized cumulative information gain, $S_{t_{j}^c,c}$ exceeds $0.2$. However, in general, it is not possible to choose integer scales $t_j^c$ so that the $S_{t_{j}^c,c}$ are exactly evenly spaced.

We emphasize that wavelet scales are learned independently for each channel $c$, allowing different scales $t_j^c$ to be assigned to each channel.
%so that we may learn different scales $t_j^c$ for each channel. 
We also note the {\methodname} procedure is easily parallelizable over graphs (for data sets consisting of multiple graphs). % Additionally, since InfoGain is performed pretraining, it can   also be used as feature selector: if a feature remains invariant over the diffusion process across graphs, it is likely not useful in distinguishing graphs. In such cases, {\methodname} will not return wavelet scales for that feature, suggesting it may be excluded.
%{\methodname} can also act as a sort of feature selector: if a feature shows invariance over the diffusion process across graphs, it is likely not useful in helping a model distinguish those graphs, {\methodname} will also not return wavelet scales, and therefore it may be considered for exclusion.

Lastly, we note that if desired, {\methodname} can furnish a simplified `average' set of custom wavelet scales, if the channel-specific scales all turn out to be similar values, or more efficient computation is desired. For example, after training the algorithm on the training data (or a subset), one can then construct a shared scales bank to use in all channels by taking the %(integer-rounded) 
median $t_j^c$ across channels $1, \ldots, C$.

\section{Experiments}\label{sec: experiments}

In order to evaluate the effectiveness of {\methodname} in improving geometric scattering-based GNNs, we build learnable scattering (LS) networks using our algorithm for wavelet scale selection and the architectural framework introduced in \cite{johnson2025manifold}. We train our `LS-{\methodnameshort}` models on six biochemical graph-classification data sets, and compare against (1) LS networks that are identical except where they use dyadic-scale diffusion wavelet filters (i.e., an ablation of {\methodname});
%as it controls for all other aspects of the LS model architecture);
and (2) a LEGS-based network.\footnote{ Code  available at \url{https://github.com/dj408/infogain}}
%First, LEGS originally had a fixed $t_J$ of 16; we generalized this so that we could run LEGS with $t_J = 32$ (to our knowledge, LEGS has never been tested at this diffusion scale).
%Consolidated results are shown in Table \ref{table:results_consolidated}, while full results can be found in Table \ref{table:results}.
Results are shown in Table \ref{table:results_consolidated}, with further details available in Table \ref{table:results}.

% In order to directly test the effect of our method, as baselines we consider two other GNNs which are otherwise identical to our method, but feature either (a) wavelets with dyadic scales (LS-dyadic) or (b) wavelets with scales learned  via a differentiable selector matrix as in \cite{tong2022learnable} (LEGS). Further model details can be found in Appendix \ref{app:models}. %This is done to directly test the effect of {\methodname}. %the novel wavelet design, the primary contribution of our work.

\subsection{Models and Data Sets}\label{app:models}

%As an example of the utility of {\methodname}, we conduct experiments with a wavelet-based GNN modeled after the network used in \cite{johnson2025manifold}.
%Each layer of this network may consist of several operations. First, a wavelet transform $\mathcal{W}_J$ (constructed using {\methodname}) is applied to each input signal $\mathbf{x}_c$. Then,  weight matrices are used to learn combinations of the filtered input signals. %\textcolor{orange}{(though for simpler models, this step is excluded)}
%%%%
% Reworded the previous sentence to includet the word ``may" and also later said full details for each data set see the appendices
%%%%%%%%%%%%%%%%%%%%%%%%%%
% PEDANTIC NOTE: We technically aren't wrong even if we see that all of the layers have all of the steps. The way MFCNs asre defined, any of the steps can predesigned or learnable. In the predefined case, you can define the combine matrix to be the identity. This was deliberately done so that our theory would apply to all of the variations. 
%%%%%
%Next, another learnable weight matrix is used to 
%and to learn combinations of the different wavelet filters. Finally, a nonlinear activation such as ReLU is applied. For full details, on the GNN used for each data set, please see Appendices \ref{app:models} and \ref{app:experiment_details}. 

% We build learnable scattering (LS) networks for graphs based on the framework introduced in \cite{johnson2025manifold}. 
Following the lead of  \cite{johnson2025manifold}, each layer in our model consists of several steps. Given an input  feature matrix, these networks first apply a filter step, followed by learnable cross-channel combinations, then learnable cross-filter combinations, and lastly, they apply nonlinear activation and reshape the layer combinations into a new  hidden feature matrix. For full details on this these steps, we refer the reader to \cite{johnson2025manifold}. 
% We abbreviate our LS networks `LS-{\methodnameshort},' and %
Here, we integrate {\methodname} into the scale selection of diffusion wavelets used in the filter step of the first layer. After the first layer in these LS models, the original input features have been recombined into a new hidden feature set; for efficiency, we do not re-apply {\methodname} to determine filter scales for hidden layers. Instead, we use the medians of the scales derived by our algorithm. 

%Inspired by \cite{johnson2025manifold}, in each layer of LS-{\methodnameshort}, we apply a wavelet transform  (designed via {\methodname}) to each of the input signals. We then use   weight matrices to learn combinations of the input channels and also to learn combinations of the different wavelet filters. 
% Because of the use of wavelets in the first step, we will think of this method as a learnable variation of the GST discussed in Section \ref{sec: background} and we refer to it as Learnable Scattering - {\methodnameshort} (LS-{\methodnameshort}). 

We note that the framework from \cite{johnson2025manifold} is designed to be flexible, as the optimal model structure will vary by data set and learning objective. Accordingly, in our experiments, the number of layers, the inclusion of the channel-combine step, and the number of output channels in each step varied by data set. However, we emphasize that these settings were the same for all methods in order to enable fair comparison. Please see Appendix \ref{app:experiment_details} for details on the specific architecture and  hyperparameters used in our experiments.
%\textcolor{orange}{Note that this architecture is deliberately (and usefully) flexible, since the optimal model structure will vary by data set and learning objective. Accordingly, in our experiments, the number of layers, the inclusion of the channel-combine step, and the number of output channels in each step varied by data set. See Section \ref{app:experiment_details} for an accounting of these parameters by data set.}

% Model training hyperparameter settings (such as learning rate, batch size, training stopping rule settings, etc.) can be found in Appendix \ref{app:experiment_details}.

%as well as final channel pooling methods, varied by data set but were kept constant between LS and LEGS models. Given the extremely large number of possible combinations of optimization hyperparameters, the full hyperparameter space was not searched beyond exploratory model fitting, and it is possible that other combinations could result in better model performance.

%\textcolor{blue}{More description of the Experiments go here. The graph classification tasks are  ``standard" so we can just cite them. The synthetic is a bit more exotic and requires some explanation. Anything else will probably need to go in the appendix. Important, Label Appendix sections using ref and refer to exactly which appendix something is in. e.g., further details on X are given in Appendix \ref{sec: Y}. Want it to be as easy as possible for the reviewer to find what they are looking for.}

%\subsection{Data sets}\label{sec:experiments:datasets}

We trained all models on five commonly-used molecular graph-classification data sets from \cite{Morris+2020}, which  were previously considered in  \cite{tong2022learnable}, as well as the Peptides-func data set from Long Range Graph Benchmarks \cite{dwivedi2022long}. For the molecular graph data sets, we utilize a 10-fold cross-validation procedure and report results as mean $\pm$ standard deviation. On Peptides-func, we use the train/validation/test sets provided in \cite{dwivedi2022long}. For details on the data sets, ablation studies, and experiment hyperparameters, see Appendices \ref{app:data sets}, \ref{app:results}, and \ref{app:experiment_details}.

%\textcolor{orange}{Described in detail in Appendix \ref{app:data sets}, we consider five commonly-used molecular graph-classification data sets with a 10-fold cross-validation (CV) procedure, plus a richer data set from the Long Range Graph Benchmark using the benchmark's train/validation/test splits \cite{dwivedi2022long}.
%For full results including ablations, see Table \ref{table:results} in Appendix \ref{app:results}. Experiment hyperparameters are in Appendix \ref{app:experiment_details}.}

% We use 10-fold cross-validation (with 80/10/10 splits) to evaluate each of the methods on five of the molecular data sets with binary graph classification tasks used in \cite{tong2022learnable}: DD \cite{dobson2003distinguishing}, PTC \cite{toivonen2003statistical}, NCI1 \cite{wale:NC1-NCI109}, MUTAG \cite{debnath1991structure}, and PROTEINS \cite{borgwardt2005protein}\footnote{We exclude ENZYMES \cite{borgwardt2005protein} since it is multi-class, and NCI109 \cite{wale:NC1-NCI109} because it is very similar to NCI1; see Appendix \ref{app:data sets} for data set descriptions.}. The mean and standard deviation accuracies and run times across 10 folds are shown in Table \ref{table:results_consolidated}. For detailed results including ablations, see Table \ref{table:results} in Appendix \ref{app:results}. Experiment hyperparameters are in Appendix \ref{app:experiment_details}.

\subsection{Results}\label{sec:experiments:results}

As shown in Table \ref{table:results_consolidated}, {\methodname} achieves the highest accuracy on four of the six data sets (PTC, NCI1, MUTAG, and Peptides-func). %PROTEINS contains only three node features, the fewest of any of the data sets. We suspect that this limited the performance of {\methodname} which lies in part in its ability to learn different scales for each feature. On DD,  {\methodname} returned approximately dyadic scales, which may explain the similar accuracy scores between all models. %However, we note that LS-{\methodname} achieved significant training speedup compared to LEGS—about twice as fast—while utilizing fewer features (since {\methodname} can identify and discard uninformative features: see Appendix \ref{app:algorithm_details}). The DD data set has 89 node features and hundreds of nodes per graph, which becomes expensive in LEGS, which must multiply a dense selector matrix with all diffusion steps on all features. In contrast, LEGS obtains the fastest training times in data sets with few node features, where the cost of these dense multiplications is less than the training costs in the LS models. 
 %We suspect that this limited {\methodname}'s performance chi
%
%Finally, PROTEINS has the fewest number of node features (three) of the data sets tested, so while {\methodname} still edged out LEGS in terms of mean accuracy, this feature set may have been too small for {\methodname}'s feature-specific scales to surpass dyadic scales. 
%In order to further understand the experimental results, we examine the scales selected by {\methodname}. 
We note that on PROTEINS, {\methodname} selects identical scales for all three node features. On DD, it finds similar, essentially dyadic scales for all features. These observations may explain the inability of {\methodname} to surpass the other models on PROTEINS, and the near-identical performance of all methods on DD.

%%%%%%%%%%%%%%%%%%%%%%%
% Small results table %
%%%%%%%%%%%%%%%%%%%%%%%
\begin{table}[htbp]
%\caption{Experimental Results}
\caption{Graph Classification Results}
\label{table:results_consolidated}
\begin{center}
% \begin{tabular}{@{}llcr@{}}
\begin{tabular}{lcr}
\toprule
\textbf{Data Set} & \textbf{Model} & \textbf{Accuracy}\\
\midrule
\multirow{3}{*}{DD \cite{dobson2003distinguishing}} 
& LS-dyadic & $\mathbf{78.11 \pm 4.70}$\\
& LS-{\methodnameshort} & $77.85 \pm 4.64$\\
& LEGS & $77.60 \pm 3.28$\\
\midrule
 \multirow{3}{*}{PTC \cite{toivonen2003statistical}} 
 & LS-{\methodnameshort} & $\mathbf{60.46 \pm 7.56}$\\
 & LEGS & $60.41 \pm 8.71$\\
 & LS-dyadic & $54.99 \pm 7.73$\\
 \midrule
 \multirow{3}{*}{NCI1 \cite{wale:NC1-NCI109}} 
 & LS-{\methodnameshort} & $\mathbf{77.47 \pm 2.42}$\\
 & LS-dyadic & $76.42 \pm 2.79$\\
 & LEGS & $69.73 \pm 1.94$\\
 \midrule
 \multirow{3}{*}{MUTAG \cite{debnath1991structure}} 
 & LS-{\methodnameshort} & $\mathbf{85.67 \pm 10.04}$\\
 & LEGS & $81.93 \pm 9.99$\\
 & LS-dyadic & $80.35 \pm 9.59$\\
\midrule
 \multirow{3}{*}{PROTEINS \cite{borgwardt2005protein}} 
 & LS-dyadic & $\mathbf{75.75 \pm 3.50}$\\
 & LS-{\methodnameshort} & $74.85 \pm 3.89$\\
 & LEGS & $74.40 \pm 4.19$\\
 \midrule
 \multirow{3}{*}{Peptides-func \cite{dwivedi2022long}}
 & LS-{\methodnameshort} & $\mathbf{81.17}$ \\
 & LS-dyadic & $78.81$ \\
 & LEGS & $77.43$ \\
\bottomrule
\end{tabular}
\end{center}
\vspace{-.5cm}
\end{table}

Our full experimental results, in which we consider several variations of each model, are shown in Table \ref{table:results} of Appendix \ref{app:results}. One notable finding from this table is how, on the Peptides-func data set (from the Long Range Graph Benchmark \cite{dwivedi2022long}), increasing the maximal scale $t_J$ from 16 to 32 boosted the performance of {\methodname} models,
% Notably, these results show that the performance of {\methodname} improves on the Peptides-func data set, taken from Long Range Graph Benchmarks, when we increase the maximal scale $t_J$ from 16 to 32. 
indicating that they may help capture long-range interactions.

\begin{figure}[htbp]
\centerline{\includegraphics[width=0.5\textwidth]{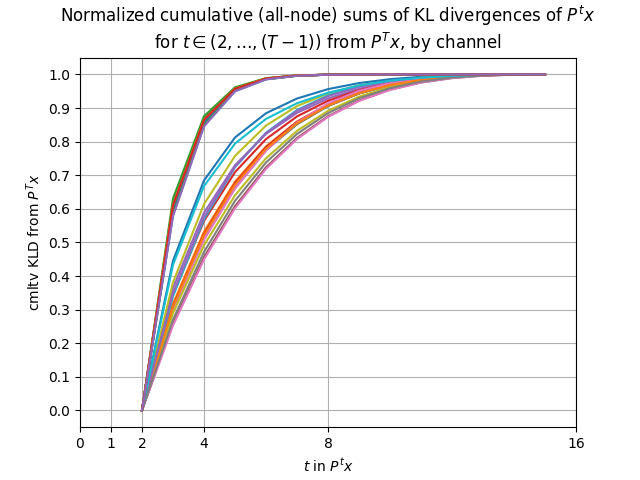}}
\caption{Diffusion “information curves” from an {\methodname} fit on the NCI1 data set. The varying steepness of these curves shows that different channels likely have different optimal diffusion wavelet scales. For reference, dyadic scales (powers of two) are marked on the $x$-axis.}
\label{fig:kld_curves_nci1}
\end{figure}

\begin{figure}[htbp]
\centerline{\includegraphics[width=0.5\textwidth]{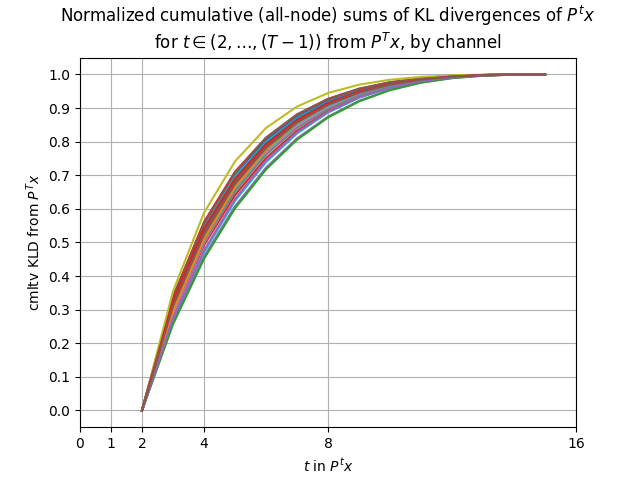}}
\caption{Diffusion “information curves” from an {\methodname} fit on the DD data set (uninformative channels excluded). The similar arcs of these curves shows that all channels likely have similar optimal diffusion wavelet scales.}
\label{fig:kld_curves_dd}
\end{figure}

% \subsubsection{Illustrations}
Finally, to illustrate the logic of our algorithm, in Figures \ref{fig:kld_curves_nci1} and \ref{fig:kld_curves_dd}, we plot the scales returned by {\methodname} for two of the data sets considered, NCI1 and DD.
%(Appendix \ref{app:algorithm_details}). 
In contrast with DD, {\methodname} is the top performing method on NCI1, and learns different sets of scales for different features.
%\textcolor{red}{In order to understand in which settings {\methodname} will be useful, we examine the output of {\methodname} on two of the data sets considered in Section \ref{sec: experiments}.}
Figure \ref{fig:kld_curves_nci1} illustrates {\methodname}'s output for the NCI1 data set. The varying steepness of the information curves suggests that different channels likely have different optimal wavelet scales. In comparison, the more consistent shape of the curves in Figure \ref{fig:kld_curves_dd} (the DD data set) suggests likely similar optimal wavelet scales across channels. Further inspection reveals these scales to be approximately dyadic. Together, these plots may help explain {\methodname} appears to boost the performance of the GNN on NCI1, but not on DD.  %\textcolor{red}{We should relate this to the fact that INFO GAIN works well on this data set, and probably `preview' this figure in the main body. DJ: done.}

%%%%%%%%%%%%%%%%%%%%%%
% KLD curves figures %
%%%%%%%%%%%%%%%%%%%%%%

\section{Conclusion}\label{sec:conclusion}
\textcolor{black}{
%Motivated by information theory, 
{\methodname} is a novel algorithm for choosing the scales for diffusion wavelets used in GSTs and GNNs. Unlike previous work \cite{tong2022learnable}, it is an unsupervised method and can learn different scales for each input channel.
%Next steps: improve efficiency of implementation, both in fitting {\methodname}, and in applying custom scales to each channel (especially for data sets with a large number of graphs). Show utility in regression and single (large) graph learning problems.
Interesting future work includes (1) further delineating which data sets benefit most from different scales for each channel, (2) incorporating {\methodname} into unsupervised wavelet-based methods \cite{venkat2024mapping, sun2024hyperedge} for data exploration, and (3) using {\methodname} to improve the design of scattering-based GNNs for combinatorial optimization problems.}

%\textcolor{orange}{Summary of where {\methodname} helps: (1) where channels of node signals have different diffusion patterns [NCI1, MUTAG vs. DD], with different optimal wavelet scales; (2) with long-range dependencies: can fine-tune diffusion filters over a long series of diffusion steps [Peptides-func results], where it's not clear how to initialize LEGS with non-dyadic scales; (3) unsupervised learning / low data}
%\textcolor{red}{Briefly discuss future directions: (1) revise algorithm to parallelize graph processing better; (2) {\methodname} likely shines on large graphs with many continuous node features (unlike the data sets tested here) (3) incorporating into other diffusion wavelet based methods both DL and non/DL ...do more experiments!}

\vfill\pagebreak

\section*{REFERENCES}
\label{sec:refs}

% List and number all bibliographical references at the end of the
% paper. The references can be numbered in alphabetic order or in
% order of appearance in the document. When referring to them in
% the text, type the corresponding reference number in square
% brackets as shown at the end of this sentence \cite{C2}. An
% additional final page (the fifth page, in most cases) is
% allowed, but must contain only references to the prior
% literature.

% Please follow the IEEE Citation Guidelines, \url{https://ieee-dataport.org/sites/default/files/analysis/27/IEEE\%20Citation\%20Guidelines.pdf} for formatting of references.

\bibliographystyle{IEEEbib}
\renewcommand{\bibsection}{} % remove auto-generated 'References' section header
\bibliography{main}

\begin{thebibliography}{10}

\bibitem{coifman:diffWavelets2006}
Ronald~R. Coifman and Mauro Maggioni,
\newblock ``Diffusion wavelets,''
\newblock {\em Applied and Computational Harmonic Analysis}, vol. 21, no. 1, pp. 53--94, 2006.

\bibitem{Mallat:2008:WTS:1525499}
St\'{e}phane Mallat,
\newblock {\em A Wavelet Tour of Signal Processing, Third Edition: The Sparse Way},
\newblock Academic Press, 3rd edition, 2008.

\bibitem{shuman:emerging2013}
David~I. Shuman, Sunil~K. Narang, Pascal Frossard, Antonio Ortega, and Pierre Vandergheynst,
\newblock ``The emerging field of signal processing on graphs: Extending high-dimensional data analysis to networks and other irregular domains,''
\newblock {\em IEEE Signal Processing Magazine}, vol. 30, no. 3, pp. 83--98, 2013.

\bibitem{Bronstein:geoDeepLearn2017}
Michael~M. Bronstein, Joan Bruna, Yann LeCun, Arthur Szlam, and Pierre Vandergheynst,
\newblock ``Geometric deep learning: Going beyond {E}uclidean data,''
\newblock {\em IEEE Signal Processing Magazine}, vol. 34, no. 4, pp. 18--42, 2017.

\bibitem{bronstein2021geometric}
Michael~M. Bronstein, Joan Bruna, Taco Cohen, and Petar Veličković,
\newblock ``{Geometric Deep Learning: Grids, Groups, Graphs, Geodesics, and Gauges},''
\newblock {\em arXiv preprint arXiv:2104.13478}, 2021.

\bibitem{gama:diffScatGraphs2018}
Fernando Gama, Alejandro Ribeiro, and Joan Bruna,
\newblock ``Diffusion scattering transforms on graphs,''
\newblock in {\em International Conference on Learning Representations}, 2018.

\bibitem{gama:stabilityGraphScat2019}
Fernando Gama, Joan Bruna, and Alejandro Ribeiro,
\newblock ``Stability of graph scattering transforms,''
\newblock in {\em Advances in Neural Information Processing Systems 33}, 2019.

\bibitem{gao:graphScat2018}
Feng Gao, Guy Wolf, and Matthew Hirn,
\newblock ``Geometric scattering for graph data analysis,''
\newblock in {\em Proceedings of the 36th International Conference on Machine Learning, PMLR}, 2019, vol.~97, pp. 2122--2131.

\bibitem{zou2020graph}
Dongmian Zou and Gilad Lerman,
\newblock ``Graph convolutional neural networks via scattering,''
\newblock {\em Applied and Computational Harmonic Analysis}, vol. 49, no. 3, pp. 1046--1074, 2020.

\bibitem{perlmutter:geoScatCompactManifold2020}
Michael Perlmutter, Feng Gao, Guy Wolf, and Matthew Hirn,
\newblock ``Geometric scattering networks on compact {R}iemannian manifolds,''
\newblock in {\em Mathematical and Scientific Machine Learning Conference}, 2020.

\bibitem{chew2022geometric}
Joyce Chew, Matthew Hirn, Smita Krishnaswamy, Deanna Needell, Michael Perlmutter, Holly Steach, Siddharth Viswanath, and Hau-Tieng Wu,
\newblock ``Geometric scattering on measure spaces,''
\newblock {\em Applied and Computational Harmonic Analysis}, vol. 70, pp. 101635, 2024.

\bibitem{bodmann2022scattering}
Bernhard~G Bodmann and Iris Emilsdottir,
\newblock ``A scattering transform for graphs based on heat semigroups, with an application for the detection of anomalies in positive time series with underlying periodicities,''
\newblock {\em Sampling Theory, Signal Processing, and Data Analysis}, vol. 22, 2024.

\bibitem{pan2020spatio}
Chao Pan, Siheng Chen, and Antonio Ortega,
\newblock ``Spatio-temporal graph scattering transform,''
\newblock {\em arXiv preprint arXiv:2012.03363}, 2020.

\bibitem{ioannidis2020efficient}
Vassilis~N Ioannidis, Siheng Chen, and Georgios~B Giannakis,
\newblock ``Efficient and stable graph scattering transforms via pruning,''
\newblock {\em IEEE Transactions on Pattern Analysis and Machine Intelligence}, vol. 44, no. 3, pp. 1232--1246, 2020.

\bibitem{mallat:scattering2012}
St{\'e}phane Mallat,
\newblock ``Group invariant scattering,''
\newblock {\em Communications on Pure and Applied Mathematics}, vol. 65, no. 10, pp. 1331--1398, October 2012.

\bibitem{czaja:timeFreqScat2017}
Wojciech Czaja and Weilin Li,
\newblock ``Analysis of time-frequency scattering transforms,''
\newblock {\em Applied and Computational Harmonic Analysis}, 2017.

\bibitem{nicola2022stability}
Fabio Nicola and S~Ivan Trapasso,
\newblock ``Stability of the scattering transform for deformations with minimal regularity,''
\newblock {\em Journal de Math{\'e}matiques Pures et Appliqu{\'e}es}, vol. 180, pp. 122--150, 2023.

\bibitem{bruna2013invariant}
Joan Bruna and St{\'e}phane Mallat,
\newblock ``Invariant scattering convolution networks,''
\newblock {\em IEEE transactions on pattern analysis and machine intelligence}, vol. 35, no. 8, pp. 1872--1886, 2013.

\bibitem{grohs:cnnCartoonFcns2016}
Philipp Grohs, Thomas Wiatowski, and Helmut B{\"o}lcskei,
\newblock ``Deep convolutional neural networks on cartoon functions,''
\newblock in {\em IEEE International Symposium on Information Theory}, 2016, pp. 1163--1167.

\bibitem{wiatowski:mathTheoryCNN2018}
Thomas Wiatowski and Helmut B{\"o}lcskei,
\newblock ``A mathematical theory of deep convolutional neural networks for feature extraction,''
\newblock {\em IEEE Transactions on Information Theory}, vol. 64, no. 3, pp. 1845--1866, 2018.

\bibitem{perlmutter:geoScatManifolds2018}
Michael Perlmutter, Guy Wolf, and Matthew Hirn,
\newblock ``Geometric scattering on manifolds,''
\newblock in {\em NeurIPS Workshop on Integration of Deep Learning Theories}, 2018,
\newblock arXiv:1812.06968.

\bibitem{perlmutter2019understanding}
Michael Perlmutter, Alexander Tong, Feng Gao, Guy Wolf, and Matthew Hirn,
\newblock ``Understanding graph neural networks with generalized geometric scattering transforms,''
\newblock {\em SIAM Journal on Mathematics of Data Science}, vol. 5, no. 4, pp. 873--898, 2023.

\bibitem{saito2017underwater}
Naoki Saito and David~S Weber,
\newblock ``Underwater object classification using scattering transform of sonar signals,''
\newblock in {\em Wavelets and Sparsity XVII}. SPIE, 2017, vol. 10394, pp. 103--115.

\bibitem{LEONARDUZZI201811}
Roberto Leonarduzzi, Haixia Liu, and Yang Wang,
\newblock ``Scattering transform and sparse linear classifiers for art authentication,''
\newblock {\em Signal Processing}, vol. 150, pp. 11--19, 2018.

\bibitem{wenkel2022overcoming}
Frederik Wenkel, Yimeng Min, Matthew Hirn, Michael Perlmutter, and Guy Wolf,
\newblock ``Overcoming oversmoothness in graph convolutional networks via hybrid scattering networks,''
\newblock {\em arXiv preprint arXiv:2201.08932}, 2022.

\bibitem{tong2022learnable}
Alexander Tong, Frederik Wenkel, Dhananjay Bhaskar, Kincaid Macdonald, Jackson Grady, Michael Perlmutter, Smita Krishnaswamy, and Guy Wolf,
\newblock ``Learnable filters for geometric scattering modules,'' 2022.

\bibitem{xu2023blis}
Charles Xu, Laney Goldman, Valentina Guo, Benjamin Hollander-Bodie, Maedee Trank-Greene, Ian Adelstein, Edward De~Brouwer, Rex Ying, Smita Krishnaswamy, and Michael Perlmutter,
\newblock ``Blis-net: Classifying and analyzing signals on graphs,''
\newblock {\em arXiv preprint arXiv:2310.17579}, 2023.

\bibitem{min2020scattering}
Yimeng Min, Frederik Wenkel, and Guy Wolf,
\newblock ``Scattering gcn: Overcoming oversmoothness in graph convolutional networks,''
\newblock in {\em Advances in Neural Information Processing Systems}, 2020, vol.~33.

\bibitem{wenkeltowards}
Frederik Wenkel, Semih Cant{\"u}rk, Stefan Horoi, Michael Perlmutter, and Guy Wolf,
\newblock ``Towards a general recipe for combinatorial optimization with multi-filter gnns,''
\newblock in {\em The Third Learning on Graphs Conference}, 2025.

\bibitem{hu2020open}
Weihua Hu, Matthias Fey, Marinka Zitnik, Yuxiao Dong, Hongyu Ren, Bowen Liu, Michele Catasta, and Jure Leskovec,
\newblock ``Open graph benchmark: Datasets for machine learning on graphs,''
\newblock {\em Advances in neural information processing systems}, vol. 33, pp. 22118--22133, 2020.

\bibitem{nt2019revisiting}
Hoang Nt and Takanori Maehara,
\newblock ``Revisiting graph neural networks: All we have is low-pass filters,''
\newblock {\em arXiv preprint arXiv:1905.09550}, 2019.

\bibitem{viswanath2024protscape}
Siddharth Viswanath, Dhananjay Bhaskar, David~R Johnson, Joao~Felipe Rocha, Egbert Castro, Jackson~D Grady, Alex~T Grigas, Michael~A Perlmutter, Corey~S O'Hern, and Smita Krishnaswamy,
\newblock ``Protscape: Mapping the landscape of protein conformations in molecular dynamics,''
\newblock {\em arXiv preprint arXiv:2410.20317}, 2024.

\bibitem{bhaskar2022molecular}
Dhananjay Bhaskar, Jackson Grady, Egbert Castro, Michael Perlmutter, and Smita Krishnaswamy,
\newblock ``Molecular graph generation via geometric scattering,''
\newblock in {\em 2022 IEEE 32nd International Workshop on Machine Learning for Signal Processing (MLSP)}. IEEE, 2022, pp. 1--6.

\bibitem{venkat2024directed}
Aarthi Venkat, Joyce Chew, Ferran~Cardoso Rodriguez, Christopher~J Tape, Michael Perlmutter, and Smita Krishnaswamy,
\newblock ``Directed scattering for knowledge graph-based cellular signaling analysis,''
\newblock in {\em ICASSP 2024-2024 IEEE International Conference on Acoustics, Speech and Signal Processing (ICASSP)}. IEEE, 2024, pp. 9761--9765.

\bibitem{viswanath2025hiponet}
Siddharth Viswanath, Hiren Madhu, Dhananjay Bhaskar, Jake Kovalic, Dave Johnson, Rex Ying, Christopher Tape, Ian Adelstein, Michael Perlmutter, and Smita Krishnaswamy,
\newblock ``Hiponet: A topology-preserving multi-view neural network for high dimensional point cloud and single-cell data,''
\newblock {\em arXiv preprint arXiv:2502.07746}, 2025.

\bibitem{chew2022manifold}
Joyce Chew, Holly Steach, Siddharth Viswanath, Hau-Tieng Wu, Matthew Hirn, Deanna Needell, Matthew~D Vesely, Smita Krishnaswamy, and Michael Perlmutter,
\newblock ``The manifold scattering transform for high-dimensional point cloud data,''
\newblock in {\em Topological, Algebraic and Geometric Learning Workshops 2022}. PMLR, 2022, pp. 67--78.

\bibitem{zou:graphScatGAN2019}
Dongmian Zou and Gilad Lerman,
\newblock ``Encoding robust representation for graph generation,''
\newblock in {\em International Joint Conference on Neural Networks}, 2019.

\bibitem{venkat2024mapping}
Aarthi Venkat, Sam Leone, Scott~E Youlten, Eric Fagerberg, John Attanasio, Nikhil~S Joshi, Michael Perlmutter, and Smita Krishnaswamy,
\newblock ``Mapping the gene space at single-cell resolution with gene signal pattern analysis,''
\newblock {\em Nature Computational Science}, vol. 4, no. 12, pp. 955--977, 2024.

\bibitem{sun2024hyperedge}
Xingzhi Sun, Charles Xu, Jo{\~a}o~F Rocha, Chen Liu, Benjamin Hollander-Bodie, Laney Goldman, Marcello DiStasio, Michael Perlmutter, and Smita Krishnaswamy,
\newblock ``Hyperedge representations with hypergraph wavelets: applications to spatial transcriptomics,''
\newblock {\em ArXiv}, pp. arXiv--2409, 2024.

\bibitem{johnson2025manifold}
David~R Johnson, Joyce~A Chew, Siddharth Viswanath, Edward~De Brouwer, Deanna Needell, Smita Krishnaswamy, and Michael Perlmutter,
\newblock ``Manifold filter-combine networks,''
\newblock {\em Sampling Theory, Signal Processing, and Data Analysis}, vol. 23, no. 2, pp. 17, 2025.

\bibitem{Morris+2020}
Christopher Morris, Nils~M. Kriege, Franka Bause, Kristian Kersting, Petra Mutzel, and Marion Neumann,
\newblock ``Tudataset: A collection of benchmark datasets for learning with graphs,''
\newblock in {\em ICML 2020 Workshop on Graph Representation Learning and Beyond (GRL+ 2020)}, 2020.

\bibitem{dwivedi2022long}
Vijay~Prakash Dwivedi, Ladislav Ramp{\'a}{\v{s}}ek, Michael Galkin, Ali Parviz, Guy Wolf, Anh~Tuan Luu, and Dominique Beaini,
\newblock ``Long range graph benchmark,''
\newblock {\em Advances in Neural Information Processing Systems}, vol. 35, pp. 22326--22340, 2022.

\bibitem{dobson2003distinguishing}
Paul~D Dobson and Andrew~J Doig,
\newblock ``Distinguishing enzyme structures from non-enzymes without alignments,''
\newblock {\em Journal of molecular biology}, vol. 330, no. 4, pp. 771--783, 2003.

\bibitem{toivonen2003statistical}
Hannu Toivonen, Ashwin Srinivasan, Ross~D King, Stefan Kramer, and Christoph Helma,
\newblock ``Statistical evaluation of the predictive toxicology challenge 2000--2001,''
\newblock {\em Bioinformatics}, vol. 19, no. 10, pp. 1183--1193, 2003.

\bibitem{wale:NC1-NCI109}
Nikil Wale, Ian~A Watson, and George Karypis,
\newblock ``Comparison of descriptor spaces for chemical compound retrieval and classification,''
\newblock {\em Knowledge and Information Systems}, vol. 14, no. 3, pp. 347--375, 2008.

\bibitem{debnath1991structure}
Asim~Kumar Debnath, Rosa~L Lopez~de Compadre, Gargi Debnath, Alan~J Shusterman, and Corwin Hansch,
\newblock ``Structure-activity relationship of mutagenic aromatic and heteroaromatic nitro compounds. correlation with molecular orbital energies and hydrophobicity,''
\newblock {\em Journal of medicinal chemistry}, vol. 34, no. 2, pp. 786--797, 1991.

\bibitem{borgwardt2005protein}
Karsten~M Borgwardt, Cheng~Soon Ong, Stefan Sch{\"o}nauer, SVN Vishwanathan, Alex~J Smola, and Hans-Peter Kriegel,
\newblock ``Protein function prediction via graph kernels,''
\newblock {\em Bioinformatics}, vol. 21, no. suppl\_1, pp. i47--i56, 2005.

\bibitem{Fey/Lenssen/2019}
Matthias Fey and Jan~E. Lenssen,
\newblock ``Fast graph representation learning with {PyTorch Geometric},''
\newblock in {\em ICLR Workshop on Representation Learning on Graphs and Manifolds}, 2019.

\bibitem{Morris2020TUDataset}
Christopher Morris, Nils~M. Kriege, Franka Bause, Kristian Kersting, Petra Mutzel, and Marion Neumann,
\newblock ``Tudataset: A collection of benchmark datasets for learning with graphs,''
\newblock in {\em ICML 2020 Workshop on Graph Representation Learning and Beyond (GRL+ 2020)}, 2020.

\bibitem{GARAVELLI2021706}
John~S. Garavelli,
\newblock ``Methods | protein data resources,''
\newblock in {\em Encyclopedia of Biological Chemistry III (Third Edition)}, Joseph Jez, Ed., pp. 706--712. Elsevier, Oxford, third edition edition, 2021.

\bibitem{singh2016satpdb}
Sandeep Singh, Kumardeep Chaudhary, Sandeep~Kumar Dhanda, Sherry Bhalla, Salman~Sadullah Usmani, Ankur Gautam, Abhishek Tuknait, Piyush Agrawal, Deepika Mathur, and Gajendra~PS Raghava,
\newblock ``Satpdb: a database of structurally annotated therapeutic peptides,''
\newblock {\em Nucleic acids research}, vol. 44, no. D1, pp. D1119--D1126, 2016.

\bibitem{loshchilov2017decoupled}
Ilya Loshchilov and Frank Hutter,
\newblock ``Decoupled weight decay regularization,''
\newblock {\em arXiv preprint arXiv:1711.05101}, 2017.

\end{thebibliography}

%%%%%%%%%%%%%%%%%%%%%%%%
%   Appendix material  %
%%%%%%%%%%%%%%%%%%%%%%%%
\appendix

% \section{Further details on the  {\methodnamecaps} Algorithm}
% \label{app:algorithm_details}

\section{Details on LEarnable Geometric Scattering (LEGS)}\label{app:legs}

%\textcolor{red}{Make sure something like this paragraph appears in the main body; DJ: it does, in the opening of the Section \ref{sec:algorithm}In Section \ref{sec:algorithm}}, we described the basic idea of the {\methodname} algorithm: To choose diffusion scales, $t_j^c$, on a channel-by-channel basis so that each wavelet coefficient $(\mathbf{P}^{{t^c_{j-1}}}- \mathbf{P}^{t^c_j})\mathbf{x}_c$ contains approximately equal amounts of information (where $\mathbf{P}$ is the lazy-random walk matrix introduced in Section \ref{sec: background} and $\mathbf{x}_c$ is the $c$-th input signal). In this section, we provide further discussion  of the background and motivation for our algorithm.

The purpose of this paper is to introduce a novel, unsupervised method for selecting diffusion wavelet scales, motivated by the idea, mentioned in Section \ref{sec: background}, that dyadic integers may be overly rigid for some applications. In this section, we review another, supervised method for selecting these scales which was introduce in \cite{tong2022learnable}. 

The scale-selection procedure from \cite{tong2022learnable} relies on an differentiable scale-selector matrix $\mathbf{F}$, which takes the form 
$$
    \mathbf{F} = \begin{bmatrix}
        &\sigma(\theta_1)_1 &\dots &\sigma(\theta_1)_{t_{\max}} \\
        &\sigma(\theta_2)_1 &\dots &\sigma(\theta_2)_{t_{\max}} \\
        &\vdots &\ddots &\vdots \\
        &\sigma(\theta_J)_1 &\dots &\sigma(\theta_J)_{t_{\max}}.
    \end{bmatrix},
    $$
    where $\sigma$ is the softmax function (applied to each row) and the $\theta_i$ are learnable parameters. 
\cite{tong2022learnable} then defines generalized diffusion wavelets by 
\begin{align}\label{eq_wavelet matrix relaxed}
    \widetilde{\Psi}_{0} \mathbf{x} &= \mathbf{x} - \sum_{t=1}^{t_{\max}} \mathbf{F}_{(1,t)} \mathbf{P}^t \mathbf{x}, \quad \widetilde{\Psi}_{J} \mathbf{x} = \sum_{t=1}^{t_{\max}} \mathbf{F}_{(J,t)} \mathbf{P}^t \mathbf{x}, \\
    %\widetilde{\mPsi}_{0} \vx &= \vx - \sum_{t=1}^m \mF_{(1,t)} \mP^t \vx, \\
    \widetilde{\Psi}_{j} \mathbf{x} &= \sum_{t=1}^{t_{\max}} \left[\mathbf{F}_{(j,t)} \mathbf{P}^t \mathbf{x} - \mathbf{F}_{(j+1,t)} \mathbf{P}^t \mathbf{x} \right], \quad 1 \leq j \leq J-1. \nonumber
\end{align}

The use of softmax is designed so that the rows of $\mathbf{F}$ are \emph{approximately} sparse, with one entry approximately equal to one and the rest approximately equal to zero.\footnote{Our implementation of LEGS is built on the original code used in \cite{tong2022learnable} (at \url{https://github.com/KrishnaswamyLab/LearnableScattering/blob/main/models/LEGS_module.py}), which omits the softmax step on rows of the selector matrix $\mathbf{F}$.}
% with one entry that is approximately one and the rest approximately zero in each row. 
Therefore, this yields 
$$
\widetilde{\Psi}_{j} \mathbf{x} \approx  \mathbf{P}^{\tilde{t}_j} \mathbf{x} - \mathbf{P}^{\tilde{t}_{j+1}} \mathbf{x},
$$
where $\tilde{t}_{j}$ corresponds to the one large entry in the $j$-th row of the selector matrix  $\mathbf{F}$.
Notably, the use of the selector matrix allows \cite{tong2022learnable} to incorporate these generalized wavelets into an end-to-end differentiable geometric scattering network, referred to as Learnable Geometric Scattering (LEGS), which they show is effective for various deep learning tasks.

\section{Data Sets}\label{app:data sets}

We first selected several molecular data sets featured in the original experiments with LEGS (cf. Table II in \cite{tong2022learnable}), focusing on those where LEGS was a top performer among the original comparisons.\footnote{
Note that in the original LEGS paper, the results presented on these data sets are from a more intensive ensembling cross-validation procedure, where 10-fold CV is performed on 80/10/10 train/valid/test splits, but each of the 10 test sets is used in combination with nine validation sets and the majority vote of nine trained models is used to calculate final accuracy scores.} Second, in order to test {\methodname} on a biological data set with more long-range relationships within graphs and rich node features (i.e., not one-hot encodings), we also selected the `Peptides-func' data set, part of the Long Range Graph Benchmark \cite{dwivedi2022long}. Table \ref{table:data set_stats} presents summary counts (number of graphs, node features, etc.) for these data sets.  We downloaded all data sets from PyTorch-Geometric's \cite{Fey/Lenssen/2019} data set library, which hosts many common data sets, including TUDataset's \cite{Morris2020TUDataset} DD, PTC, NCI1, and MUTAG, and PROTEINS sets. 

%%%%%%%%%%%%%%%%%%
% data sets stats %
%%%%%%%%%%%%%%%%%%
\begin{table}[htbp]
\caption{Data Sets Summary Counts}
\label{table:data set_stats}
\label{table:bioinformatics_data sets}
\begin{center}
\begin{tabular}{lrrrr}
\toprule
\textbf{Data Set} & \textbf{Graphs} & \makecell{\textbf{Node} \\ \textbf{Features}} & \makecell{\textbf{Avg.} \\ \textbf{Nodes}} & \makecell{\textbf{Avg.} \\ \textbf{Edges}} \\
\midrule
DD & 1,178 & 89 & 284.32 & 715.66 \\
PTC & 344 & 18 & 14.29 & 14.69 \\
NCI1 & 4,110 & 37 & 29.87 & 32.30 \\
MUTAG & 188 & 7 & 17.93 & 19.79 \\
PROTEINS & 1,113 & 3 & 39.06 & 72.82 \\
% \midrule
Peptides-func & 15,535 & 9 & 150.94 & 153.65 \\
\bottomrule
\end{tabular}
\end{center}
\end{table}

\textbf{DD} \cite{dobson2003distinguishing} is compromised of 1178 graphs with 89 one-hot encoded features encoding structurally unique proteins from the Protein Data Bank \cite{GARAVELLI2021706}, labeled as enzymes (41.3\%) or non-enzymes (58.7\%). %The data set, which we downloaded from PyTorch-Geometric, is the TUDataset version \cite{Morris2020TUDataset}.
In cross-validation, {\methodname} dropped varying numbers of uninformative features, of which 30 were most common, thus leaving 59 features in data used to fit the `subset' models for DD.
%We used moments pooling, batch size 128, learning rate 10^{-3}, class-bal loss, burn-in 100, patience 50
%MFCN layers: None-(8, )
%{\methodname} quantiles: (0.125, 0.25, 0.375, 0.5, 0.625, 0.75, 0.875); 
% Given the larger size and large proportion of sparse features of data set, as an additional experiment, instead of re-running {\methodname} before each cross-validation fold, we instead first ran {\methodname} on a random 50\% of all graphs, in order to obtain the `uninformative' features to drop and then obtain the average (median) scales for the remaining features (`LS-{\methodname}-16-50\%-drop-med'). The original data set contains 89 node features, 54 of which we excluded after {\methodname}'s feature selection. For comparison, we fit LEGS and LS-dyadic models on both the full original feature set and on this 35-feature subset.
We note that the {\methodname} model achieves comparable mean accuracy at almost a three-fold speedup per epoch compared to LEGS models. % We also fit an {\methodname} model that runs the algorithm de novo with each fold's training set to identify uninformative features to drop (`LS-{\methodname}-16-drop'), and then uses wavelet scales (potentially) unique to each channel. 
However, the LS-dyadic models performed just as well, which we suspect is due to the fact that {\methodname} returns nearly dyadic scales for this data set.

\textbf{PTC} \cite{toivonen2003statistical} contains 344 graphs of chemical compounds labeled as carcinogenic or not to rats. Each graph %(again, in the TUdata set version as downloaded from PyTorch Geometric) 
has 18 node features (one-hot atom type). The data set also includes edge features (bond type), which we did not use. For this data set, for {\methodname} models, we fit the algorithm on the training data of each fold, and automatically dropped uninformative features (which ranged from two to six). The data set is slightly imbalanced, with 44.2\% positive samples. Since our intention with the`LS-dyadic-[$t_J$]' models is to ablate the wavelet scales returned by {\methodname} versus dyadic scales, we also fit these models (with suffix \mbox{`-subset'} in the results table) on a trimmed feature set, that is, excluding the two most commonly dropped features (across CV folds) by {\methodname} (features 0 and 13, indexing from 0). 
%We used moments pooling, batch size 32, learning rate 10^{-3}, class-bal loss, burn-in 100, patience 50
%MFCN layers: (8, 4)-(8, 4) 
%{\methodname}: quantiles: (0.25, 0.5, 0.75), uninformative feature strategy: drop.

\textbf{NCI1} \cite{wale:NC1-NCI109} contains 4110 graphs of chemical compounds labeled by whether the compound showed activity in inhibiting the growth of non-small cell lung cancer cell lines. Node features are one-hot encoded atoms (37 types), and edges represent whether two atoms are share a bond. The target classes are approximately balanced. Models with the suffix \mbox{`-subset'} used a trimmed feature set excluding features 12, 30, and 36 (indexing from 0), representing the most commonly dropped features by {\methodname} during cross-validation.
% From the original 37 atom types (node features), we removed 18 rare (uninformative) types using {\methodname}'s feature selection.
%We used moments pooling, batch size 256, learning rate 10^{-2}, class-bal loss, burn-in 100, patience 32
%MFCN layers: (8, 4)-(8, 4) 
%{\methodname} quantiles: (0.125, 0.25, 0.375, 0.5, 0.625, 0.75, 0.875) 

\textbf{MUTAG} \cite{debnath1991structure} is a collection of 188 graphs of nitroaromatic compounds divided into two classes based on their mutagenicity (ability to cause genetic mutations, as a likely carcinogen) based on the Ames test with the bacterial species \textit{Salmonella typhimurium}. Nodes are one-hot encoded atoms (seven types); edges represent bonds (with corresponding edge labels of four, one-hot encoded bond types; not used). The data set is imbalanced approximately 2:1, with 66.5\% positive class samples. In {\methodname} models' CV runs, the only feature dropped was at index 4; hence the `subset' runs of other models exclude only this feature.

\textbf{PROTEINS} \cite{dobson2003distinguishing, borgwardt2005protein} contains 1178 graphs of proteins labeled as enzymes (40.4\%) or non-enzymes (59.6\%), with three node features only: one-hot encodings of helices, sheets or turns. Edges encode that two nodes are either (a) neighbors along an amino acid sequence, or (b) one of three nearest neighbors in Euclidean space within the protein structure. Note that the {\methodname} algorithm did not find any uninformative features; hence, no other models were fit with a data subset as an ablation. 

\textbf{Peptides-func} is  Long Range Graph Benchmark \cite{dwivedi2022long} data set which contains 15,535 graphs of peptides from SATPdb \cite{singh2016satpdb}, pre-split into stratified train, validation, and test sets (70\%, 15\%, and 15\%). Peptides are short chains of amino acids, which lend themselves to graphs with more nodes and larger diameters than small molecules, but with similar average node degree (under similar featurization, i.e., using heavy atoms as node features); hence peptides are a good candidate for learning long-range graph dependencies in medium-sized graphs \cite{dwivedi2022long}. This data set originally has 10 classification targets, labeling various peptide functions (antibacterial, antiviral, etc.), though with highly variable class imbalances. However, we simplified it to a binary classification data set by taking the single most well-balanced target (62.7\% positive class labels). Notably, these graphs have nine rich (not one-hot encoded) node features from OGB molecular featurization of their molecular SMILES code \cite{hu2020open}, of which {\methodname} dropped one as uninformative (feature 5).

\section{Extended Results}\label{app:results}

We present further experimental results in Table \ref{table:results}, expanding on those shown in Table \ref{table:results_consolidated}. Please note the following:
\begin{itemize}

\item The `-drop' suffix on {\methodname} models indicates that, in each CV fold, any uninformative features identified were dropped from the feature set. %{\methodname} models that drop uninformative features are marked with the suffix \mbox{`-drop,'} while those 

\item Models that set their scales to the median of informative features' scales are marked with \mbox{`-med'}.
%The suffix \mbox{`-subset'} marks comparison models trained with a feature set reduced by the features most commonly dropped by {\methodname} across folds.
%We mark {\methodname} models that drop uninformative features with the suffix \mbox{`-drop'}, and those that set the scales of uninformative features to the median of informative features' scales with `{-med}'. 

\item Importantly, as another ablation, when {\methodname} trims uninformative features in cross-validation folds, we also trim its most commonly dropped features from the data set when training LEGS models. These models are labeled with the \mbox{`-subset'} suffix in Table \ref{table:results}. The features dropped in these subset models are recorded for each data set in Appendix \ref{app:data sets}. In contrast, \mbox{`-full'} indicates the full, original feature set was used. 

\item As a reference baseline, we also include the results of the best-scoring LEGS model as reported in \cite{tong2022learnable} (LEGS-16-original).\footnote{In \cite{tong2022learnable}, results are reported for several variations of LEGS architectures, including use of a support vector machine (with radial basis function) as the classifier head, or a two-layer fully-connected network (FCN) head (with or without an attention layer between LEGS and FCN modules). Note that we copy the best score from \cite{tong2022learnable}, regardless of which LEGS model achieved it, and with the caveat that these models were trained under different hyperparameters, GPU hardware, and (ensembling) CV experimental design (and timing results are not reported).}
%Table \ref{table:results} also includes the top LEGS model by accuracy score reported in Table II of \cite{tong2022learnable} (LEGS-16-Original-[$\cdot$]), though these models were trained under different hyperparameters, GPU hardware, and (ensembling) CV experimental design (and timing results are not reported).

\item Since the Peptides-func data set was introduced as an example of where long-range interactions are important, as part of Long Range Graph Benchmarks \cite{dwivedi2022long}, on this data set, we also compared using a max diffusion step of $t_J = 16$ versus $t_J = 32$.\footnote{Note that this leads to four versus five wavelet filters in LEGS, given its dyadic initialization, whereas the number of wavelet filters (and selector quantiles for them) in {\methodname} is a tunable hyperparameter.} Therefore, in results for this data set, %results were more mixed between the $t_j = 16$ and $t_j = 32$ models. T
the \mbox{`-16'} suffix  represents a maximum diffusion step $t_J = 16$ was used, and \mbox{`-32'} represents $t_j = 32$. However, in exploratory modeling with the other five data sets, i.e., those featured in \cite{tong2022learnable}, $t_j = 32$ consistently performed similarly or worse for all models compared to using $t_J = 16$. Thus for clarity, these results are excluded for these models.

\end{itemize}

To facilitate comparisons between the shortened results in presented in  Table \ref{table:results_consolidated} and the extended results presented in  Table \ref{table:results}, we note that methods in Table \ref{table:results_consolidated},  LS-{\methodname}, LEGS, and LS-dyadic, correspond to LS-{\methodname}-16-drop, LEGS-16-full, and LS-dyadic-16-full shown in Table \ref{table:results}. The sole exception is Peptides-func: since this data set was designed to highlight the importance of long-range interactions, we chose either the `16' or `32' variation of each method, whichever performed best. We choose these variations to display in Section \ref{sec: experiments} since they were generally the best performing variation of each method. Additionally, we note that the variations of LEGS and LS-dyadic with \mbox{`-drop'} are included for completeness. However, they are, in some sense, not true baselines, since they are constructed using information obtained via {\methodname}.

%In our experiments, we fix the maximum diffusion step $t_J = 16$ for all models.\footnote{In exploratory modeling with these data sets using $t_j = 32$ consistently performed similarly or worse for all models compared to using $t_J = 16$.}

Overall, the results shown in Table \ref{table:results} expand on those in Table \ref{table:results_consolidated} and discussed in the main text. LS-{\methodname} fails to surpass the baselines on PROTEINS and DD, but is the top performing method on the other four data sets. These results suggest that {\methodname} may be less useful where: (1) the optimal wavelet scales are already dyadic; or (2) the data set has few features, and especially when all features show similar diffusion patterns. On the other hand, our results show that {\methodname} is likely more useful when the data set has: (1) features with heterogeneous, non-dyadic diffusion patterns, and/or (2)  long-range dependencies.

To illustrate the latter point, we note that on the Peptides-func data set, the {\methodname} models with $t_J = 32$ learned best, indicating that {\methodname} may be well-suited for modeling long-range dependencies efficiently. We hypothesize that {\methodname} differentiated itself with an ability to partition a long, multi-channel diffusion process with a small set of non-overlapping, informative bandpass wavelets, making learning complex patterns on such graphs more tractable for a neural network.

% We also point out that {\methodname}-drop models were generally the top performer even with reduced feature sets (resulting from {\methodname}'s built-in feature selection), versus the LS-dyadic and LEGS \mbox{`-full'} models.
% This is perhaps explained in part by reducing `learning noise' through discarding uninformative features. LS-{\methodname}-32-drop beat LS-{\methodname}-32-med (which set the scales for the one uninformative feature to the median scales of the others) on Peptides-func.
% \textcolor{blue}{I think we should delete this paragraph. We don't display results on {\methodname}-full. [It is probably true, but likely does more harm than good unless we compare {\methodname} drop VS full], which we should do in future iterations. Comment this text rather than deleting it so we have it for then]}\textcolor{purple}{same applies to the next paragraph. I am sure this is true. But we cant see it from your tables.}

Finally, we note that our initial modeling efforts also suggested some lessons on tuning {\methodname}'s hyperparameters: (1) that the zero replacement strategy (whether we replace the zero values of $\mathbf{q}_c^t$ with a small constant such as $10^{-2}$, or use the linear halfway point between zero and the minimal nonzero value, i.e., $\frac{1}{2}\min\{\mathbf{q}_c^t:\mathbf{q}_c^t>0\}$) can have a large effect on the scales returned, especially with sparse features such as one-hot encodings of atom types; (2) that larger $t_J$ is not always better, and may be counterproductive to learning if the graph structures do not contain many long-range dependencies; and (3) increasing the number of wavelets (i.e., with a larger number of selection quantiles) is similarly not always better, if the data doesn't support it (note that {\methodname} models learned best on PTC and MUTAG with a small set of wavelets).

% Finally, we observe that LEGS models from our implementation achieved higher accuracy scores than those (from an ensembled, majority-vote classification procedure) recorded in Table II of \cite{tong2022learnable}. That is, except for on MUTAG (in which a LEGS model utilizing an attention layer; something we do not test here) and NCI1, which scored higher than the LEGS model of our implementation. We suppose that the hyperparameter and pooling choices of our experiments contributed to these observed performance differences.

%\textcolor{orange}{Reiterate that {\methodname} found uninformative channels on a random sample of 50\% of all DD graphs; for other data sets, {\methodname} was re-run on the full train data for each CV fold; the most commonly dropped features across folds were excluded from the data sets used to train the \mbox{`-subset'} ablation models.}

%\textcolor{orange}{As an ablation on {\methodname}'s automatic feature selection (as opposed to its wavelet scale selection), we train LEGS-16 models ($t_J$ is fixed at 16 in \cite{tong2022learnable}) with both subsetted and full feature sets (these are suffixed \mbox{`-subset'} versus \mbox{`-full'} in the results table).}

\section{Experimental details}
\label{app:experiment_details}

\subsection{{\methodname} Hyperparameters}
{\methodname} is a flexible, tunable technique, dependent on several hyperparameters. The full list of hyperparameters is: 
\begin{itemize}
    \item the maximum diffusion step $t_J$;
    \item the number of wavelet filters, which also corresponds to the number information quantiles;
    \item the strategy employed to handle uninformative features: for example, these may be dropped, or replaced by the median scales of the informative channels, or replaced by (zero-padded) dyadic scales;
    \item the proportion of graphs or nodes in the training set fed to the {\methodname} algorithm: a smaller proportion may be used to speed up the estimation of wavelet scales (but may decrease the optimality of wavelet scales in small or noisy data sets);
    \item the batch size when fitting the algorithm in batches of multiple graphs: a smaller batch size may lead to less parallel processing of graphs and a longer fit time;
    \item the KL divergences summed across graphs (within channels) can also be re-weighted for class imbalance in graph classification problems;
    \item the strategy or constant used to replace zeros in features may be modified (since KL divergence uses logarithms, zeros \textit{must} be replaced, and the strategy selected here can have a substantial effect on the wavelet scales returned by {\methodname}).
\end{itemize}

\textcolor{black}{
Finally, note that if the number of the number of scales $t_j^c$ is too large (or the KL curve of a particular channel is  `too steep'), duplicate scales will be returned by the {\methodname} Algorithm. That is, we will have $t_{j+1}^c=t_j^c$ for some $j$'s. This will imply that $j$-th wavelet filter $\mathbf{P}^{t_{j+1}^c} - \mathbf{P}^{t_{j}^c}$ will be equal to zero. In our current implementation, we leave these zero features in for the sake of simplicity. However, if desired, one could also drop the zero features. Alternatively, one could also consider replacing the wavelet coefficient with a low-pass at that scale, i.e., include $\mathbf{P}^{t_j^c}\mathbf{x}_c$ as an output of the generalized wavelet transform. We leave further exploration of this idea to future work. 
}
% \textcolor{red}{
% One could also consider introducing a (non-wavelet) operator consisting of $\mathbf{P}^t$ alone, where $t$ is the duplicate diffusion step returned by the {\methodname} algorithm that bridges multiple selector quantiles. (However, we have yet to test this approach.)
% }

The {\methodname}-related hyperparameter values used in our experiments are collected in Table \ref{table:method_hyperparams}. Note that values for `Quantiles Interval' in Table \ref{table:method_hyperparams} reflect the quantile stride (i.e., proportion of information gain desired in each wavelet produced by {\methodname}). For example, a value of $1/8$ reflects quantiles of $(0.125, 0.25, 0.5, ..., 0.875)$. The `Zeros Sub.' entries indicate the value substituted for any zeros in the normalized probability vectors in the {\methodname} algorithm (which KL divergence cannot process). The `LS Model Layers' entries summarize the LS model architecture used in {\methodname} models and their dyadic LS ablations: the first tuple represents the output number of hidden features in the cross-channel combination step for each layer (or `None' for not using this step); the second tuple stores the same for the cross-filter combination step. For instance, (8, 4)-(8, 4) encodes two layers where the cross-channel and cross-filter steps produce eight hidden features in the first layer, and four in the second layer.

%%%%%%%%%%%%%%%%%%%%%%%%%
% hyperparameters table %
%%%%%%%%%%%%%%%%%%%%%%%%%
\setlength{\tabcolsep}{2pt}

\begin{table}[htbp]
\caption{{\methodname} Hyperparameters by Data Set}
\label{table:method_hyperparams}
\begin{minipage}{0.5\textwidth}
\begin{center}
\begin{tabular}{lcccc}
\toprule
\textbf{Data Set} & \makecell{\textbf{Quantile} \\ \textbf{Interval}} & \makecell{\textbf{Class-Bal.} \\ \textbf{KLDs}} & \makecell{\textbf{Zeros} \\ \textbf{Sub.}} & \makecell{\textbf{LS Model} \\ \textbf{Layers}} \\
\midrule
DD & $1/8$ & yes & $10^{-2}$ & None-(8,) \\
PTC & $1/4$ & no & $10^{-2}$ & (8,4)-(8,4)  \\
NCI1 & $1/8$ & yes & $10^{-2}$ & (8,4)-(8,4) \\
MUTAG & $1/4$ & no & $10^{-2}$ & (8,)-(8,) \\
PROTEINS & $1/5$ & no & $10^{-2}$ & (8,4)-(8,4) \\
Peptides-func & $1/8$ & no & $\frac{1}{2}\mathrm{min(nz)}$
\footnote{More precisely: $\frac{1}{2}\mathrm{min} \{\mathbf{q}^{t}_{c} : \mathbf{q}^{t}_{c} > 0\}$ (`nz' is nonzero)} & (8,4)-(8,4) \\
\bottomrule
\end{tabular}
\end{center}
\end{minipage}
\end{table}

\begin{table}[htbp]
\caption{Training Hyperparameters by Data Set}
\label{table:experiment_params}
\begin{center}
\begin{tabular}{lcccccc}
\toprule
\textbf{Data Set} & \makecell{\textbf{Validate} \\ \textbf{Every}} & \makecell{\textbf{Patience}} & \makecell{\textbf{Pooling}} & \makecell{\textbf{Batch} \\ \textbf{Size}} & \makecell{\textbf{Learn} \\ \textbf{Rate}} \\
\midrule
DD  & 1 & 50 & moments & 64 & $10^{-3}$ \\
PTC  & 1 & 50 & moments & 32 & $10^{-3}$ \\
NCI1  & 1 & 32 & moments & 256 & $10^{-2}$ \\
MUTAG  & 1 & 100 & moments & 16 & $0.005$ \\
PROTEINS  & 1 & 50 & moments & 128 & $10^{-3}$ \\
Peptides-func  & 5 & 50 & max + mean & 512 & $0.005$ \\
\bottomrule
\end{tabular}
\end{center}
\end{table}

\subsection{Training Hyperparameters}

Training hyperparameters shared across models for each data set Table \ref{table:experiment_params}.\footnote{Because a full search of the complete hyperparameter space is unfeasible, experiment hyperparameters were set largely by exploratory modeling. Hence it is possible that other parameter settings may give different (or better) results. However, the experiments were fair in the sense that shared hyperparameters were kept constant across models, and were tuned not just with attention to {\methodname} models, but all models tested. For instance, a batch size of 128 for the DD data set resulted in an out-of-memory error for LEGS, so all models were re-run on DD with a batch size of 64.}
For all models, the classifier head is a five-layer fully-connected network with layers of 128, 64, 32, 16, and one perceptrons; we also used batch normalization layers and ReLU activations. All models and experiments utilize the AdamW optimizer \cite{loshchilov2017decoupled} and a cross-entropy loss function, which was re-weighted for class imbalance except in the NCI1 and MUTAG datasets, where doing so hindered learning. 

Training proceeded until at least 100 `burn-in' epochs were reached, and then the `patience' number of epochs (where no decrease in validation set loss was achieved, checked every or every fifth epoch, depending on the data set) was also reached; note that `patience' epochs could overlap with `burn-in' epochs. Timing results reflect training on a single NVIDIA RTX 2000 Ada 16 GB  GPU (except for LEGS results copied from Table II of \cite{tong2022learnable}, which does not report timing results).

Finally, for fair comparison with LEGS, we made several changes to the original LEGS code in our implementation, which generally improved its performance. That is, we swapped LEGS's channel pooling method of normalized (statistical) moments for unnormalized moments (which reduced computation time and appeared to improve accuracy). We also added additional pooling capabilities (such as mean, max, etc.), and abstracted the code so $t_J$, previously fixed at 16, could be any positive power of two. %\textcolor{blue}{MP: Dave, does your code feature the option to turn softmax on/off?}

%%%%%%%%%%%%%%%%%%%%%%%%%
%   full results table  %
%%%%%%%%%%%%%%%%%%%%%%%%%
% \clearpage
% \noindent
% \onecolumn
\begin{table*}[t] % table* = table span ignores cols
\centering
\begin{minipage}[t]{\textwidth} % [t][10cm][t]
\small 
% \begin{minipage}{0.8\textwidth}
% \begin{table*}[htbp]
\caption{Full Experimental Results}
\label{table:results}
% centerline centers the minipage
% \centerline{
% \hline
\begin{center}
% \begin{tabular}{|c|c|c|c|}
% \resizebox{\columnwidth}{!}{%
% these 2 lines force the table to use the full width
\setlength\tabcolsep{0pt}
\begin{tabular*}{\linewidth}{@{\extracolsep{\fill}} lcrrr}
% \begin{tabular}{lcrrr}
\toprule
\textbf{Model} & \textbf{Accuracy} & \textbf{Sec. per epoch} & \textbf{Num. epochs} & \textbf{Min. per fold}\footnote{\label{table:results:fn:minperfold} Excludes {\methodname} algorithm execution time, which is approximately fixed by fold (at a given batch size and proportion of the data used to fit the algorithm), whereas folds' total training times can vary widely, given stochastic optimization and random training data combinations in shuffled batch training.} \\
\midrule
\textit{DD}\footnote{\label{table:results:fn:tvtsplits}10-fold CV, 80/10/10 train/valid/test splits.} \\
\midrule
LS-dyadic-16-full & $\mathbf{78.11 \pm 4.70}$ & $0.63 \pm 0.02$ & $101 \pm 58$ & $1.07 \pm 0.61$ \\
LS-dyadic-16-subset & $78.02 \pm 4.53$ & $0.63 \pm 0.02$ & $94 \pm 55$ & $0.99 \pm 0.57$ \\
LEGS-16-subset & $78.02 \pm 5.52$ & $1.99 \pm 0.07$ & $86 \pm 45$ & $2.85 \pm 1.54$ \\
LS-{\methodname}-16-drop & $77.85 \pm 4.64$ & $0.72 \pm 0.03$ & $109 \pm 48$ & $1.31 \pm 0.57$ \\
LEGS-16-full & $77.60 \pm 3.28$ & $1.98 \pm 0.06$ & $76 \pm 37$ & $2.51 \pm 1.21$ \\
LEGS-16-original-RBF\footnote{\label{table:results:fn:tongensemblingcv}10-fold CV ensembling procedure from \cite{tong2022learnable}. `RBF' denotes that the model's classifier head is a support vector machine with radial basis function kernel. `FCN' denotes a fully-connected network classifier head. `ATTN-FCN' denotes an attention layer before the FCN head.} & $72.58 \pm 3.35$ & - & - & - \\
\midrule
\textit{PTC}\footref{table:results:fn:tvtsplits} \\
\midrule
LS-{\methodname}-16-drop & $\mathbf{ 60.46 \pm 7.56}$ & $ 0.17 \pm 0.01 $ & $ 48 \pm 33 $ & $ 0.13 \pm 0.09 $ \\
% LS-{\methodname}-32-apuc-drop & $ 60.43 \pm 8.57 $ & $ 0.22 \pm 0.01 $ & $ 30 \pm 29 $ & $ 0.11 \pm 0.11 $ \\
LEGS-16-full & $ 60.41 \pm 8.71 $ & $ 0.07 \pm 0.01 $ & $ 68 \pm 54 $ & $ 0.08 \pm 0.06 $ \\
% LEGS-32-full & $ 60.41 \pm 9.20 $ & $ 0.10 \pm 0.01 $ & $ 70 \pm 29 $ & $ 0.11 \pm 0.05 $ \\
LS-dyadic-16-subset & $ 57.84 \pm 8.68 $ & $ 0.15 \pm 0.01 $ & $ 61 \pm 42 $ & $ 0.16 \pm 0.11 $ \\
LEGS-16-original-RBF\footref{table:results:fn:tongensemblingcv} & $57.26 \pm 5.54$ & - & - & - \\
LEGS-16-subset & $ 56.39 \pm 6.95 $ & $ 0.07 \pm 0.01 $ & $ 84 \pm 40 $ & $ 0.10 \pm 0.05 $ \\
% LS-dyadic-32-subset-2 & $ 56.09 \pm 7.16 $ & $ 0.21 \pm 0.02 $ & $ 30 \pm 38 $ & $ 0.11 \pm 0.14 $ \\
LS-dyadic-16-full & $ 54.99 \pm 7.73 $ & $ 0.16 \pm 0.01 $ & $ 84 \pm 69 $ & $ 0.22 \pm 0.18 $ \\
% LS-dyadic-32-full & $ 54.33 \pm 8.99 $ & $ 0.21 \pm 0.02 $ & $ 57 \pm 46 $ & $ 0.20 \pm 0.16 $ \\
% LS-dyadic-32-subset & $\mathbf{ 61.03 \pm 9.24}$ & $ 0.21 \pm 0.02 $ & $ 42 \pm 38 $ & $ 0.15 \pm 0.14 $ \\
% LS-{\methodname}-16-drop & $ 60.46 \pm 7.56 $ & $ 0.17 \pm 0.01 $ & $ 48 \pm 33 $ & $ 0.13 \pm 0.09 $ \\
% LS-{\methodname}-32-drop & $ 60.43 \pm 8.57 $ & $ 0.22 \pm 0.01 $ & $ 30 \pm 29 $ & $== 0.11 \pm 0.11 $ \\
% LEGS-16-full & $ 60.41 \pm 8.71 $ & $ 0.07 \pm 0.01 $ & $ 68 \pm 54 $ & $ 0.08 \pm 0.06 $ \\
% LEGS-32-full & $ 60.41 \pm 9.20 $ & $ 0.10 \pm 0.01 $ & $ 70 \pm 29 $ & $ 0.11 \pm 0.05 $ \\
% LS-dyadic-16-subset & $ 57.84 \pm 8.68 $ & $ 0.15 \pm 0.01 $ & $ 61 \pm 42 $ & $ 0.16 \pm 0.11 $ \\
\midrule
\textit{NCI1}\footref{table:results:fn:tvtsplits} \\
\midrule
LS-{\methodname}-16-drop & $\mathbf{77.47 \pm 2.42}$ & $ 1.31 \pm 0.05 $ & $ 135 \pm 36 $ & $ 2.96 \pm 0.74 $ \\
LS-dyadic-16-full & $ 76.42 \pm 2.79 $ & $ 1.23 \pm 0.06 $ & $ 103 \pm 33 $ & $ 2.12 \pm 0.70 $ \\
LS-dyadic-16-subset & $ 76.20 \pm 2.32 $ & $ 1.22 \pm 0.04 $ & $ 107 \pm 42 $ & $ 2.18 \pm 0.83 $ \\
LEGS-16-original-RBF\footref{table:results:fn:tongensemblingcv} & $74.26 \pm 1.53$ & - & - & - \\
LEGS-16-full & $ 69.73 \pm 1.94 $ & $ 0.50 \pm 0.03 $ & $ 92 \pm 59 $ & $ 0.77 \pm 0.50 $ \\
LEGS-16-subset & $ 68.44 \pm 2.80 $ & $ 0.50 \pm 0.03 $ & $ 52 \pm 37 $ & $ 0.44 \pm 0.31 $ \\
% LS-{\methodname}-16 & $\mathbf{ 75.74 \pm 1.96}$ & $ 1.30 \pm 0.04 $ & $ 108 \pm 31 $ & $ 2.34 \pm 0.70 $ \\
% LS-{\methodname}-32 & $ 75.62 \pm 2.19 $ & $ 1.80 \pm 0.06 $ & $ 104 \pm 44 $ & $ 3.12 \pm 1.29 $ \\
% LS-dyadic-32 & $ 75.52 \pm 2.85 $ & $ 1.45 \pm 0.05 $ & $ 102 \pm 36 $ & $ 2.47 \pm 0.86 $ \\
% LS-dyadic-16 & $ 75.40 \pm 2.85 $ & $ 1.18 \pm 0.03 $ & $ 107 \pm 35 $ & $ 2.10 \pm 0.67 $ \\
% LEGS-16-subset & $ 70.75 \pm 2.65 $ & $ 0.45 \pm 0.03 $ & $ 82 \pm 41 $ & $ 0.62 \pm 0.33 $ \\
% LEGS-32-subset & $ 65.91 \pm 2.40 $ & $ 0.60 \pm 0.03 $ & $ 52 \pm 23 $ & $ 0.52 \pm 0.23 $ \\
\midrule
\textit{MUTAG}\footref{table:results:fn:tvtsplits} \\
\midrule
LS-{\methodname}-16-drop & $\mathbf{85.67 \pm 10.04}$ & $ 0.09 \pm 0.01 $ & $ 94 \pm 75 $ & $ 0.15 \pm 0.12 $ \\
LEGS-16-original-ATTN-FCN\footref{table:results:fn:tongensemblingcv} & $84.60 \pm 6.13$ & - & - & - \\
LEGS-16-subset & $ 82.40 \pm 9.12 $ & $ 0.06 \pm 0.01 $ & $ 83 \pm 52 $ & $ 0.08 \pm 0.05 $ \\
LEGS-16-full & $ 81.93 \pm 9.99 $ & $ 0.06 \pm 0.01 $ & $ 81 \pm 53 $ & $ 0.08 \pm 0.05 $ \\
LS-dyadic-16-full & $ 80.35 \pm 9.59 $ & $ 0.10 \pm 0.01 $ & $ 134 \pm 84 $ & $ 0.21 \pm 0.14 $ \\
LS-dyadic-subset & $ 77.28 \pm 18.50 $ & $ 0.09 \pm 0.01 $ & $ 67 \pm 81 $ & $ 0.10 \pm 0.12 $ \\
\midrule
\textit{PROTEINS}\footref{table:results:fn:tvtsplits}\footnote{\label{table:results:fn:nouninformativefeats}No uninformative features were found by {\methodname}, so no `subset' models were necessary.} \\
\midrule
LS-dyadic-16-full & $\mathbf{75.75 \pm 3.50}$ & $0.40 \pm 0.02$ & $81 \pm 36$ & $0.55 \pm 0.23$ \\
LS-{\methodname}-16-drop & $74.85 \pm 3.89$ & $0.42 \pm 0.02$ & $64 \pm 27$ & $0.45 \pm 0.19$ \\
LEGS-16-full & $74.40 \pm 4.19$ & $0.14 \pm 0.02$ & $54 \pm 42$ & $0.13 \pm 0.10$ \\
LEGS-16-original-FCN\footref{table:results:fn:tongensemblingcv} & $71.06 \pm 3.17$ & - & - & - \\
\midrule
\textit{{Peptides-func}}\footnote{One training run, on the 70/15/15 splits supplied by the benchmark data set (hence no $\pm$ st. dev. from cross-validation, and `Min. per fold' reflects total training time).} \\
\midrule
LS-{\methodname}-32-drop & $\mathbf{81.17}$ & $5.60 \pm 0.06$ & $350$ & $32.66$ \\
LS-{\methodname}-32-med & $80.44$ & $5.54 \pm 0.07$ & $270$ & $24.95$ \\
LS-{\methodname}-16-med & $79.58$ & $4.59 \pm 0.07$ & $145$ & $11.10$ \\
LS-dyadic-16-full & $78.81$ & $3.81 \pm 0.06$ & $190$ & $12.08$ \\
LS-{\methodname}-16-drop & $78.42$ & $5.58 \pm 0.06$ & $160$ & $14.89$ \\
LS-dyadic-32-full & $78.04$ & $4.94 \pm 0.09$ & $115$ & $9.47$ \\
LS-dyadic-32-subset & $77.86$ & $5.03 \pm 0.06$ & $145$ & $12.15$ \\
LS-dyadic-16-subset & $77.73$ & $3.95 \pm 0.08$ & $160$ & $10.53$ \\
LEGS-16-full & $77.43$ & $1.64 \pm 0.05$ & $130$ & $3.55$ \\
LEGS-32-full & $76.10$ & $2.50 \pm 0.03$ & $160$ & $6.67$ \\
LEGS-32-subset & $76.10$ & $2.33 \pm 0.04$ & $160$ & $6.21$ \\
LEGS-16-subset & $75.12$ & $1.51 \pm 0.05$ & $80$ & $2.01$ \\
\bottomrule
% \footref{table:results:fn:tvtsplits} 10-fold CV, 80/10/10 train/valid/test splits. \\
%Accuracies are those achieved on each fold's test set. \\
% $^{\mathrm{b}}$ 10-fold CV, 80/20 splits (no test set). \\
%Accuracies are those achieved on each fold's validation set; the small data set size leads to highly unstable results when using a test set. \\
% $^{\mathrm{c}}$ 10-fold CV ensembling procedure from \cite{tong2022learnable}. \\
% $^{\mathrm{d}}$ Batch size of 128 led to GPU out-of-memory error. Trained with batch size 64 instead. \\
% $^{\mathrm{*}}$ Used entire original feature set. \\
\end{tabular*}
% } % resizebox closure
\end{center}
% \vspace{-6em}
% } % centerline closure
\end{minipage}
\end{table*}
% \twocolumn

\end{document}